\documentclass[10pt,twocolumn,letterpaper]{article}

\usepackage{cvpr}
\usepackage{times}
\usepackage{epsfig}
\usepackage{graphicx}
\usepackage{amsmath}
\usepackage{amssymb}
\usepackage{algorithm}
\usepackage{algorithmic}

\usepackage[pagebackref=true,breaklinks=true,letterpaper=true,colorlinks,bookmarks=false]{hyperref}

 \cvprfinalcopy 

\ifcvprfinal\fi
\begin{document}

\title{Registration and Fusion of Multi-Spectral Images Using a Novel Edge Descriptor}

\author{Nati Ofir\\
\and
Shai Silberstein\\
\and
Dani Rozenbaum \and
Yosi Keller \and
Sharon Duvdevani Bar
}

\maketitle
\begin{abstract}
	In this paper we introduce a fully end-to-end approach for multi-spectral image registration and fusion. Our method for fusion combines images from different spectral channels into a single fused image by different approaches for low and high frequency signals. A prerequisite of fusion is a stage of geometric alignment between the spectral bands, commonly referred to as registration. Unfortunately, common methods for image registration of a single spectral channel do not yield reasonable results on images from different modalities. For that end, we introduce a new algorithm for multi-spectral image registration, based on a novel edge descriptor of feature points. Our method achieves an accurate alignment of a level that allows us to further fuse the images. As our experiments show, we produce a high quality of multi-spectral image registration and fusion under many challenging scenarios.
\end{abstract}

\section{Introduction}

\begin{figure}
	\centering
	\includegraphics[width=80px]{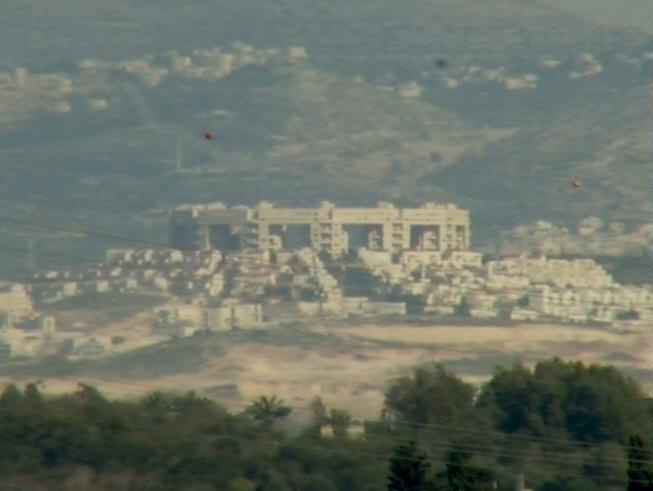}~
	\includegraphics[width=80px]{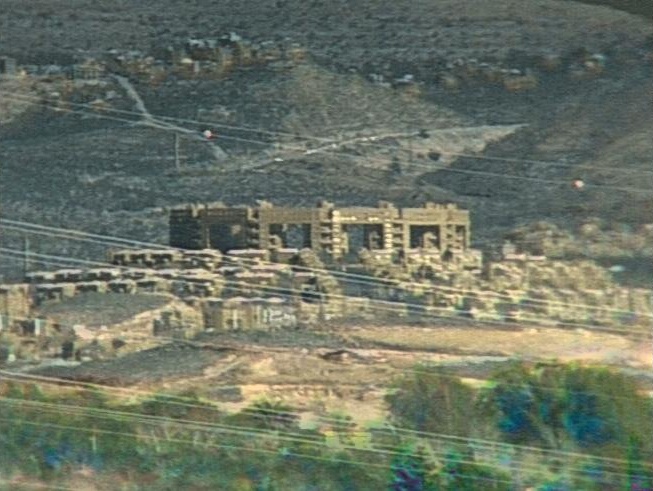}~
	\includegraphics[width=80px]{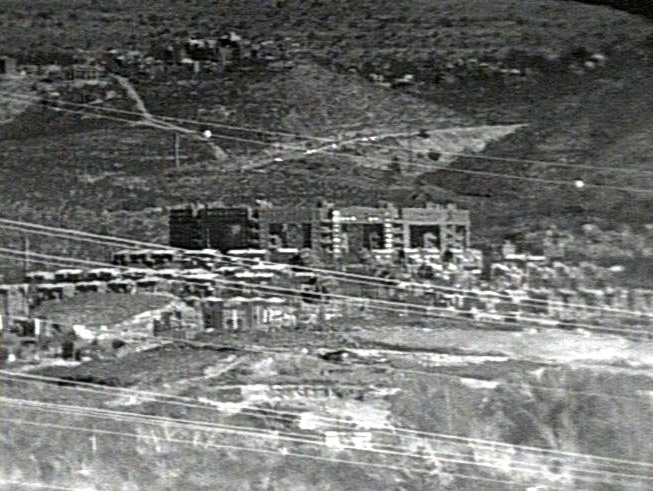}
	\caption{Example of a multi-spectral image registration and fusion. Left: RGB image of the visible channel $0.4-0.7\mu m$, contains the information about the color. Right: aligned grayscale MWIR images $3-5\mu m$, "sees" behind the haze, captures information about the cars in the far road. Center: fused image with the advantages of both channels, color information combines with good visibility of far objects.}
	\label{fig:1}       
\end{figure}

This paper addresses the problem of multi-spectral registration and fusion. Different spectral channels capture different scenes, therefore, their fusion is interesting, and their registration is challenging. See Figure \ref{fig:1} for example, the visible channel in wavelength $0.4-0.7 \mu m$, captures the colors of the scene on the one hand, but suffers from low visibility due to haze on the other hand. On the right, the Middle-Wave-Infrared (MWIR) channel in wavelength of $3-5 \mu m$, is complementary to the visible channel. It does not capture colors, but utilizes a good visibility even in the presence of haze. For an example of good visibility, see the objects in the far road, where "hot" cars can be easily seen in the thermal (MWIR) image. The fused image in the middle contains the advantages of both channels: good visibility of far distant objects combined with color information. In this work we introduce a method for performing registration of this kind of images, that manages to align multi-spectral images even though their captured information is highly different. In addition, we introduce a fast method for fusion of color-visible with IR-thermal images.

Image registration is a famous problem in computer vision with profound research. A classic way to align two images from the same modal \cite{SIFT_Registration,zitovaSurvey} is by Scale-Invariant-Feature-Transform (SIFT) \cite{SIFT}. Even though this method is robust, it does not reproduce its success on images from different modalities. The reason for this failure is that the appearance of the same object, under different spectral wavelengths, varies. Since we found out that unfortunately, SIFT descriptors are not invariant to different wavelengths, we offer to perform registration of multi-spectral images in a similar way, but to use instead a novel edge descriptor. Our method detects keypoints by Harris corner detection \cite{harris} and describes them using an edge descriptor. We measure the keypoint similarities by correlation of descriptors and we compute the final transformation by a new variant of Random-Sample-Consensus (RANSAC). As our experiments show, this new approach produces accurate registration results on multi-spectral images, since their edge information is indeed invariant to different wavelengths. Moreover, since measuring similarities by our edge descriptors is fast, our scheme can be incorporated into real-time vision systems.

The paper is organized as follows. In Section \ref{sec:previous} we cover previous work on the topics of registration and fusion. In Section \ref{sec:registration} we introduce our method, for registration of multi-spectral images, based on our new edge descriptor. Then, in Section \ref{sec:fusion}, we explain how to fuse two registered images: a color image of the visible channel, and a MWIR gray-scale image. In Section \ref{sec:experimetns}, we demonstrate evaluations of the accuracy of our registration algorithm, as well as examples of fusion of images captured from different sensors.

\section{Previous Work} \label{sec:previous}

Image registration is a fundamental problem in computer vision which has been studied for decades. Image registration methods \cite{survey,zitovaSurvey} are being used to fuse images, for matching stereo to recover object shape, or to produce a wide panorama image. Early methods rely on primitive characteristics of the images, for example, techniques that find translation by correlation \cite{correlation} of image pixels, \cite{lucas1981iterative} solves registration based on the spatial intensity gradient of the image. \cite{reddy1996fft} solves translation, rotation and scale by and FFT-based technique. More advanced methods rely on key-points \cite{harris,SIFT} and invariant descriptors to find feature-based registration \cite{SIFT_Registration}.

A distinct group of works tried to solve the problem of image registration on images from different modalities or spectral channels \cite{chen2015sirf, shen2014multi}. In \cite{mutualInformation} registration was carried out by two medical images based on maximization of mutual information. Since contours and gradients are invariant to different spectral images, \cite{multiSpectralSIFT,irani1998robust,keller2006multisensor,li1995contour} offer to utilize them for registration. When trying to solve only translation, measuring correlation on Sobel image \cite{sobel}, or Canny image \cite{Canny} may help to register multi-modal images. We show experimentally that our method outperforms these approaches. Note that we align images of a wide spectral range, visible to MWIR, while other methods \cite{multiSpectralSIFT,keller2006multisensor} work on a narrower range, visible to Near-Infrared (NIR). When spectral ranges are closer, the captured images are more similar and therefore easier to align. \cite{lghd2015} solves registration by FAST features \cite{takacs2010unified} and unique descriptors for non linear intensity variations. \cite{aguilera2016learning} measures cross spectral similarity by Convolutional-Neural-Network (CNN). They achieve high accuracy in classification of patches as same or different, however, their method cannot be incorporated into a fully registration scheme due to insensitivity to slight geometrical transformations. Our approach utilizes the fact that edges are invariant to multi-spectral images, and use it within a feature-based registration. Edge descriptors are effective and relevant for our purpose also because they are sensitive to geometric transformation and since their computation is fast. In Section \ref{sec:registration} we describe in details our novel edge descriptor and how to use it for multi-spectral registration.

Fusion of images retrieved from different sensors is a fundamental image processing problem addressed by many previous works \cite{huang2014spatial,dong2015high}. The most basic group of fusion methods, named $\alpha$ blending, and can be written as a linear combination, $\alpha I_1+(1-\alpha)I_2$,
where $I_1,I_2$ are the input images. Early methods of image fusion are based on wavelets transform \cite{wavelets}, or on Laplacian pyramid of an image \cite{laplacian}. More advanced approaches rely on Principal-Component-Analysis (PCA) transform of the fused images \cite{pca}, and on Intensity-Hue-Saturation (IHS) technique \cite{IHS}. In this work, we introduce a new technique named High-Pass-Low-Pass (HPLP). This method applies different fusion techniques to the low frequency signals and the high frequencies, it is described in details in Section \ref{sec:fusion}.

\section{Multi-Spectral Image Registration} \label{sec:registration}

The algorithm for multi-spectral image registration consists of 3 major steps: corners detection by Harris \cite{harris}, feature matching by a novel edge descriptor and iterative Random-Sample-Consensus (RANSAC) \cite{ransac} for outliers rejection.

\textbf{Corners Detection.} Our multi-spectral registration scheme is based on feature-points, whereas each point is a Harris corner \cite{harris}. This method is useful for our purpose, since the groups of corners of the two spectral images has a large overlap. We apply the following algorithm to detect the corners in each spectral image.


Let $I$ be the input image, $I_x,I_y$ be the horizontal and vertical derivatives of $I$ respectively. Given a Gaussian window $w$ compute the Harris matrix for every pixel:
\begin{equation} \label{eq:harrisMatrix}
A = \sum_u \sum_v w(u,v)
\begin{pmatrix}
I_x(u,v)^2 & I_x(u,v)I_y(u,v) \\ 
I_x(u,v)I_y(u,v) & I_x(u,v)^2
\end{pmatrix}
\end{equation}
A corner is characterized by a large positive eigenvalues $\lambda_1,\lambda_2$ of the matrix $A$. Since the computation of exact eigenvalues is expensive, we instead compute the corner score:
$S = \lambda_1\lambda_2-k(\lambda_1+\lambda_2)^2 = \det(A)-k\cdot trace^2(A)$,
where $k$ is a sensitivity parameter. For every pixel $x$ we compute the score $S$ to obtain a Harris score image $S(x)$. In practice, we apply a non-maximal-suppression by a window of size $w_1\times w_1$, such that every corner pixel, is a local maximum in its surrounding window in $S(x)$.

\textbf{Feature Matching by an Edge Descriptor.} We describe every corner point in one of the multi-spectral images by a novel edge descriptor. The uniqueness of this descriptor is its invariance over different wavelengths. Note that although the same object is acquired differently in the visible channel and in the MWIR channel, it has high similarity after applying our edge descriptor. 


Denote by $C_V$ the set of corners in the visible image and by $C_{IR}$ the corners in the MWIR image. Every point $p\in C_V$ or $p\in C_{IR}$ is basically described by its center pixel $(x_p,y_p)$. In addition, in a surrounding window of size $w_2\times w_2$, we compute the Canny \cite{Canny} edge image $E_p$ and the Gradient directions $G_p$. Note that $E_p$ is a binary image, $G_p$ is an image with indexes of directions and reasonable results are achieved also with smaller size of window. The whole $360^o$ directions are quantized in $G_p$ to $k_1$ bins. To conclude, the descriptor $D_p$ of a point contains the following information:
$D_p = \{x_p,y_p,E_p,G_p\}$.

Given two points, $p \in C_V$ and $q\in C_{IR}$, we match them according to their descriptors $D_p,D_q$. We first require that the difference in the gradient direction, will be less or equal than a single bin in any pixel $x$:
$SameGrad_{p,q}(x) =  |G_p(x)-G_q(x)| mod 16 \le 1$.
Hence, the similarity measure between $p$ and $q$ is the normalized correlation of the pixels that are in the same Gradient directions:
\begin{equation} \label{eq:similarity}
Similarity(p,q) = \frac{\sum_x E_p(x) \wedge E_q(x) \wedge SameGrad_{p,q}(x)}{\sqrt{\sum_x E_q(x)}}.
\end{equation}
Then, for every corner $p\in C_V$, we iterate over all the corners $q\in C_{IR}$ and search for the match with the highest similarity. We normalize $Similarity(p,q)$ by $\sqrt{\sum_x E_q(x)}$ to avoid high scores in cases of many edge pixels in $E_q$.

\textbf{Iterative RANSAC.} Eventually, we compute the final transformation between the images by our iterative version of RANSAC \cite{ransac}. We do so in 3 iterations, the first one is a by applying RANSAC in the regular way as follows. We compute for each corner $p\in C_V$ its best match $q^*\in C_{IR}$, and thus we get a group of matches $M_1$. This group contains inlier and outlier matches, therefore we apply the RANSAC outliers rejection approach. We sample small groups of matches $m_1,...,m_l,... \in M_1$ and compute by least square the transformation $T_l$ that best describes these matches for every subgroup $m_l$. Each match induces two linear constraints and therefore if our transformation $T$ is characterized by $n$ parameters, $|m_l| = \lceil \frac{n}{2} \rceil$. Finally, we select the best transformation $T^*$ derived from the best group of matches $m^*$, that has the greatest support in the whole group of matches $M_1$. The support of a transformation is the number of other matches that "agree" with it, a match "agree" with a transformation if
\begin{equation} \label{eq:ransacDistance}
||T_{2\times3}\begin{pmatrix}
x_p \\ y_p \\ 1
\end{pmatrix}-
\begin{pmatrix}
x_{q^*} \\ y_{q^*} \\ 1
\end{pmatrix}
||_2 \le RansacDistance.
\end{equation}
The $RansacDistance$ in the first iteration is $rd_1$. We denote by $T_1$ the transformation that is found by RANSAC in the first iteration.

In the second iteration we repeat the same process as in the first, but with initial guess $T_1$. We use this initial guess to restrict the group of all matches in the second iteration $M_2$ as follows. $p\in C_V$ can be matched to $q\in C_{IR}$ only if their distance under $T_1$ (similar to Eq. \eqref{eq:ransacDistance}), is less than $MatchDistance$. In iteration 2, $MatchDistance = md_1$ and $RansacDistance=rd_1$ such that $md_1<\infty$. We denote by $T_2$ the transformation found by our iterative RANSAC in the second iteration.

In the third and final iteration, we use $T_2$ as initial guess and we decrease the thresholds, such that $RansacDistance = rd_2$ and $MatchDistance = md_2$ such that $rd_2<rd_1,md_2<md_1$. Because these thresholds are relatively low, the final transformation is more accurate. Finally, we denote by $T_3$ the transformation found by RANSAC in the third iteration. This is the final transformation of our multi-spectral registration algorithm. Note that we can search for different types of transformations between the input images. Our implementation supports searching for any type of geometric relations.

\section{Fast Image Fusion} \label{sec:fusion}

Given two registered images, a visible channel image $V$ and a MWIR image $IR$, we would like to obtain their colored fusion image $F_c$, which its luminance gray-scale image is $F$. Since $V$ is an RGB image, we denote by $Y(V)$ its luminance channel and by $R(V),G(V),B(V)$ its color channels. In this section we introduce a new fusion technique which we call High-Pass-Low-Pass (HPLP). This method fuses $Y(V)$ with $IR$ by different approach for the low-frequency signals in the image and for the high-frequency signals. Then we explain how to restore the colors of $V$ in the fused color image $F_c$. Note that our fusion method is fast and can be run in real-time algorithms easily.

Given an image $I$ and a Gaussian kernel $g_\sigma$ with standard deviation of $\sigma$, we divide the image to high and low frequencies by a simple convolution,
$LP_\sigma(I) = I*g_\sigma$
for low pass, and then the high-pass is
$HP_\sigma(I) = I-LP_\sigma(I)$.
We fuse the low-frequencies by alpha-fusion,
\begin{equation} \label{eq:alphaLowPass}
LP_\sigma(F) = \alpha\cdot LP_\sigma(Y(V))+(1-\alpha)\cdot LP_\sigma(IR).
\end{equation}
The high-frequency image is fused by maximum criterion, such that in each pixel we take the value of the channel with the highest absolute value. Denote by $b$ the indicator 
$b = 1_{|HP_\sigma(Y(V))|\ge |HP_\sigma(IR)|}$,
then in any pixel,
\begin{equation}
HP_\sigma(F) = b\cdot HP_\sigma(Y(V))+(1-b)\cdot HP_\sigma(IR).
\end{equation}
Now, the fused image is the low-pass plus the high-pass times gain to emphasize sharpness,
$F_\sigma = LP_\sigma(F)+gain \times HP_\sigma(F)$.
We repeat the whole process for three levels of $\sigma = \sigma_1,\sigma_2,\sigma_3$ and then we fuse all the three with equal weights,
$F = \frac{1}{3}(F_{\sigma_1}+F_{\sigma_2}+F_{\sigma_3})$.

Now that we have a gray-scale fusion of $V$ and $IR$, we would like to compute the color fusion $F_c$. For that end we apply color restoration from $V$. Our criterion is that the following ratio will be preserved,
$\frac{F}{Y(V)} = \frac{R(F_c)}{R(V)} = \frac{G(F_c)}{G(V)} = \frac{B(F_c)}{B(V)}$.
Therefore, the final color fusion is this ratio times the input visible image,
$F_c = \frac{F}{Y(V)} \cdot V$.

\section{Experiments} \label{sec:experimetns}

We tested our algorithm for multi-spectral registration and fusion on various real images captured by multi-spectral imaging systems. The transformation between the channels is unknown in advance and we found it automatically by our method to create fusion examples. We evaluated the accuracy of our approach by creating a dataset of manually-aligned images, and then finding a solution to a known simulated transformation. We tested our method on this dataset and compared it to several approaches for solving simple multi-spectral registrations. Our registration code is written in C++ and its runtime is less than 4 seconds on a CPU. By utilizing parallel computing or GPU its runtime could be dramatically decreased, even to the level of real time algorithm. Our fusion scheme is implemented in Matlab and its runtime is negligible since it contains simple arithmetic computations.

In Table \ref{fig:3} we present a comparison between our accuracy to those of correlation of Canny \cite{Canny} images, correlation of Sobel \cite{sobel} images and maximization of mutual-information \cite{mutualInformation}. We also compare to \cite{lghd2015} but this method achieves no accurate results on our dataset. We simulated a known translation between the channels and tested the various methods when trying to solve translation only. Eventually, we find the Euclidean distance between the known translation to the output of each method to derive accuracy in pixels. We repeat this test for several images and store the mean of accuracy of all iterations. We test the performance for visible to Short-Wave-Infrared (SWIR, $1.4-3 \mu m$) registration, and for visible to MWIR registration.
It can be seen that in both wavelengths we achieve the most accurate results among all the compared approaches. Moreover, our algorithm is the only one that produces registration with sub-pixel accuracy (its score is less that 1 pixel in both channels).

\begin{figure}
	\begin{tabular}{c c}
		\includegraphics[width=100px]{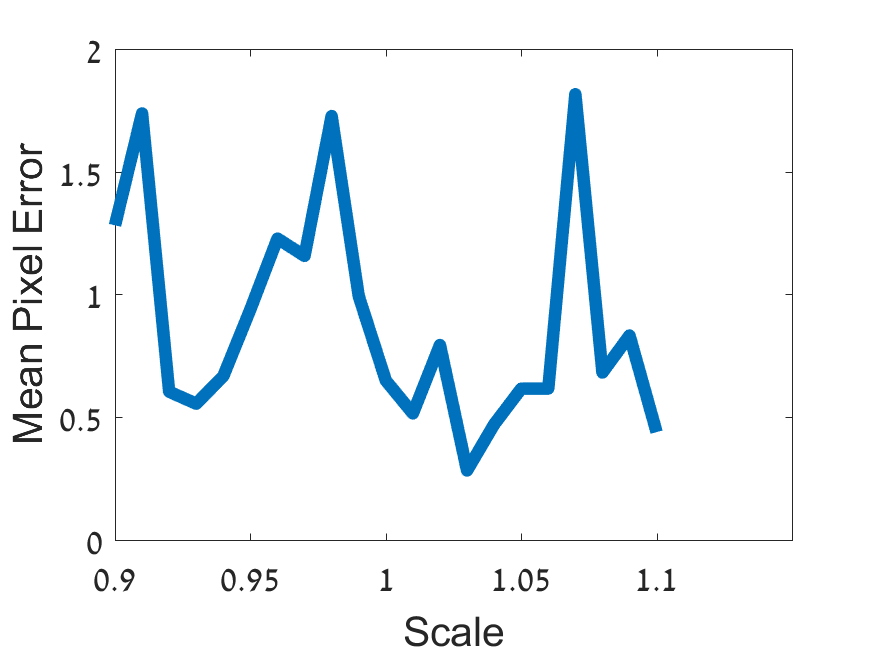}
		\includegraphics[width=100px]{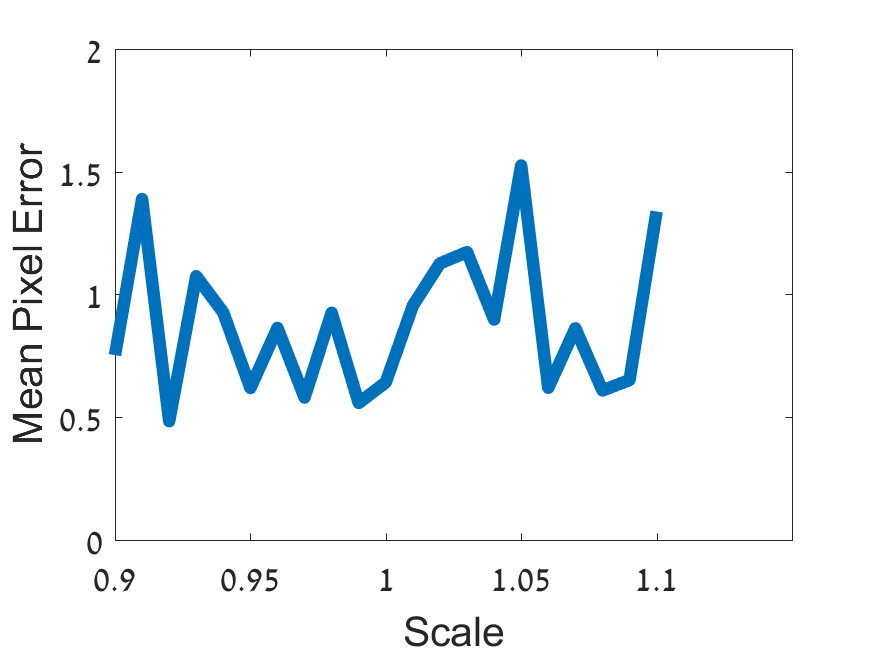}
	\end{tabular}
	\caption{Accuracy in pixels of multi-spectral registration as a function of scale between the images. Left: visible to SWIR registration. Right: visible to MWIR registration. The accuracy is around 1 pixel for all scales between 0.9 to 1.1 in both graphs.}
	\label{fig:4}
\end{figure}

We evaluate our accuracy on different scales between two spectral channels in Figure \ref{fig:4}. We report the Euclidean distance between the translation we found automatically to the one we simulated. The same distance in the scale parameter is less than $0.001$ and therefore it is negligible. It can be seen that our accuracy is around 1 pixel, in both channels SWIR and MWIR. In addition, we maintain this level of accuracy along all simulated scale transformations from $0.9$ to $1.1$.

\begin{table}
	\begin{tabular}{ l | c | c } \label{table}
		Algorithm & VIS-SWIR & VIS-MWIR \\
		\hline
		Our method & \textbf{0.62} & \textbf{0.76} \\
		Canny & 2.13 & 1.43 \\
		Sobel & 3.84 & 3.2 \\
		Mutual Information & 1.38 & 2.48 \\
		LGHD & 24.1 & 8.13
	\end{tabular}
	\caption{Accuracy in pixels of multi-spectral registration when searching for translation only. Our method is compared to correlation of Canny, correlation of Sobel, maximization of mutual-information and LGHD. As can be seen (in bold), our approach achieves the highest accuracy and it is the only one which is sub-pixel accurate (score is less than 1).}
	\label{fig:3}
\end{table}

\begin{figure}
	\begin{tabular}{c}
		\includegraphics[width=70px]{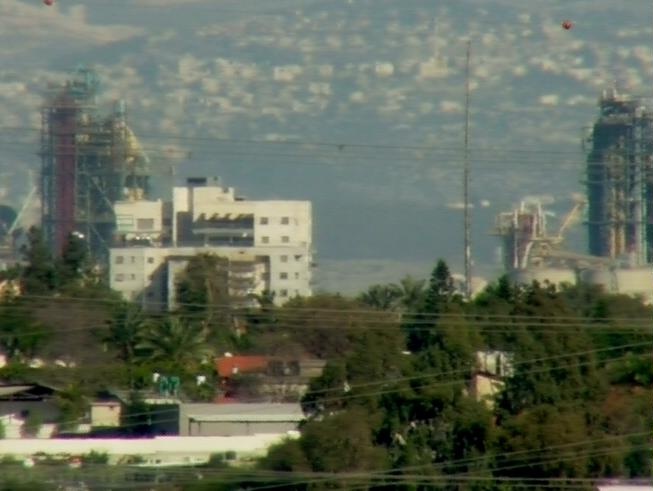}~\includegraphics[width=70px]{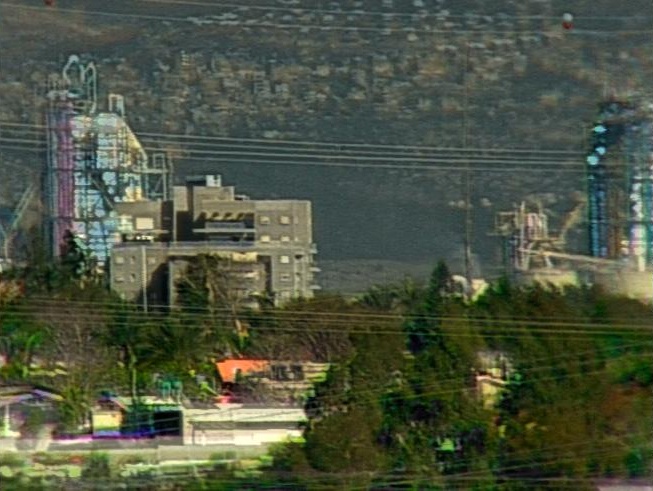}~\includegraphics[width=70px]{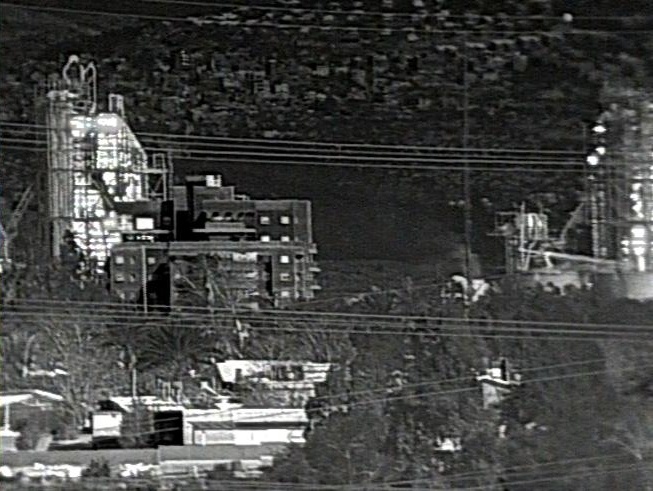}
	\end{tabular}
	\caption{Example of a multi-spectral image registration and fusion. Left: RGB image of the visible channel $0.4-0.7\mu m$, contains color information. Right: aligned grayscale MWIR images $3-5\mu m$ captures thermal information of objects. Center: fusion image with the advantages of both channels.}
	\label{fig:5}
\end{figure}

We show an interesting example of registration and fusion in Figure \ref{fig:5} . We perform registration between different spectral channels visible and MWIR, by our method. Then, we fuse the aligned images by HPLP technique. It can be seen that the left column contains all the information about colors of the captured objects, while the right column contains all the thermal information of the objects. The image of the "hot" factory building demonstrates how the two channels are naturally different. This object is very hot in the MWIR image and yet very colorful in the visible image. In the fused image we can see its colors together with its brightness indicating a high temperature.

\begin{figure}
	\begin{tabular}{c}
		\includegraphics[width=70px]{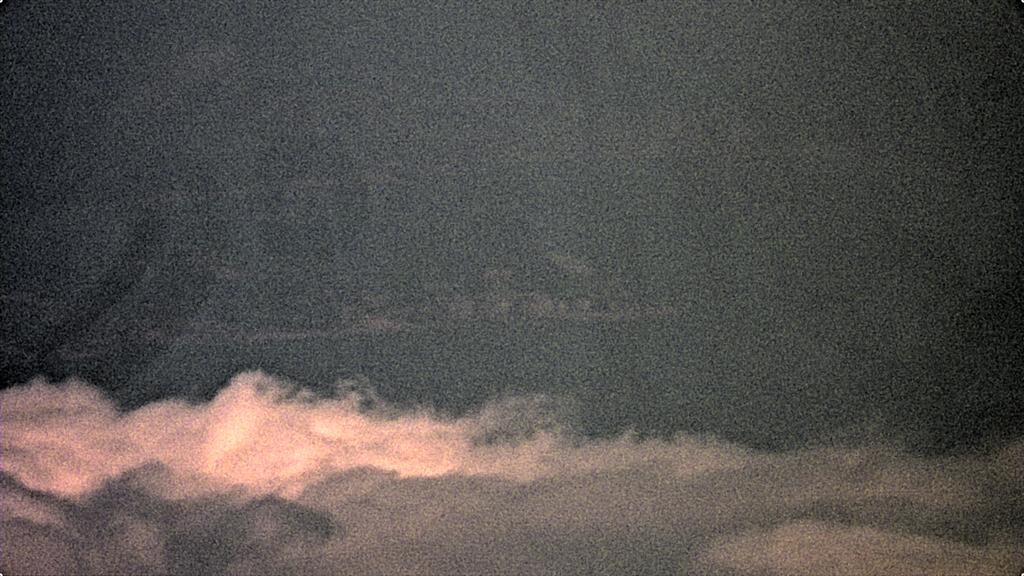}~\includegraphics[width=70px]{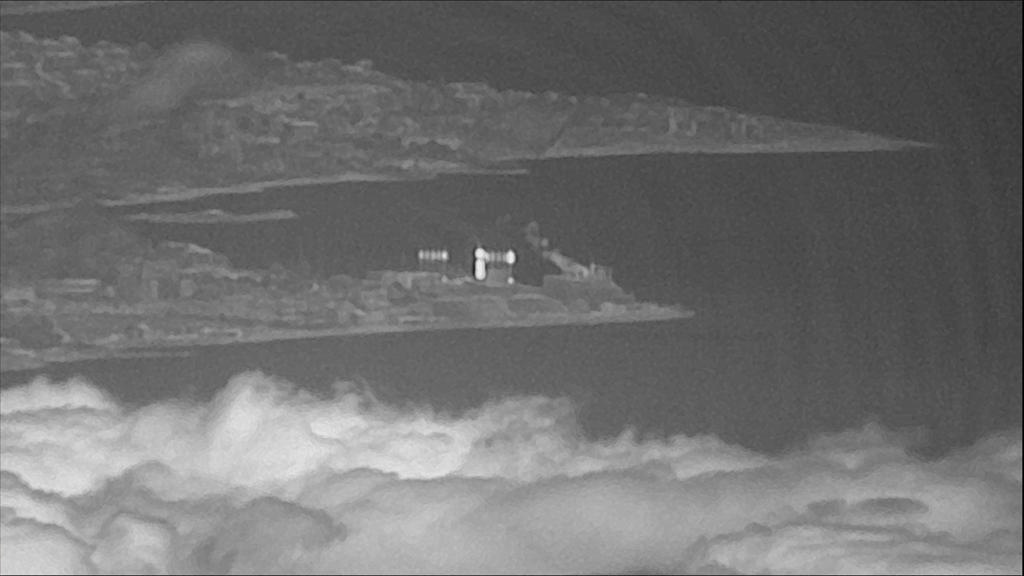}~\includegraphics[width=70px]{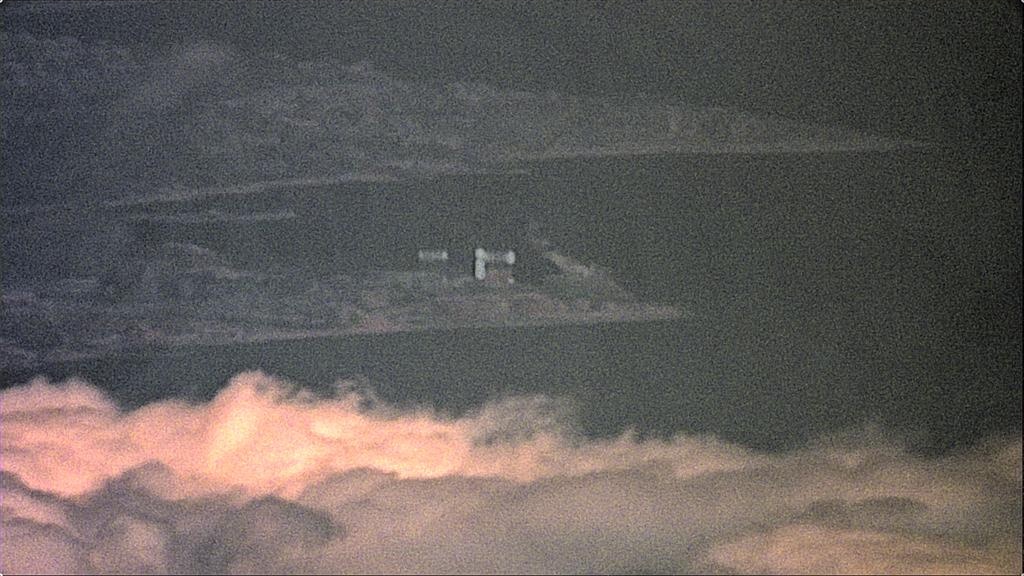}
	\end{tabular}
	\caption{Example of HPLP fusion captured by an aerial platform. From left to right: visible image, MWIR image and regular HPLP fusion.}
	\label{fig:6}
\end{figure}

In Figure \ref{fig:6} we show images captured by and aerial platform. The images provide an interesting example of fusion. The visible image captures color of the clouds, but cannot provide information beyond far distances, whereas the MWIR image captures the far land. As can be seen, the fusion method contains the interesting information captured in both channels.




\section{Conclusions} \label{sec:conclusions}

We introduced a novel method to register images captured by multi-spectral sensors. Our approach is feature based and it utilizes the characteristic that edge descriptors are invariant to different wavelengths. The multi-spectral registration is accurate enough to allow the fusion of images. For that end we developed a new technique for fusion, named High-Pass-Low-Pass, that fuses the images in different ways for the low and high frequencies. Our method outputs informative fusion results that use the benefits of both spectral channels.

{\small
	\bibliographystyle{ieee}
	\bibliography{../egbib}

\begin{thebibliography}{10}\itemsep=-1pt

\bibitem{lghd2015}
C.~Aguilera, A.~D. Sappa, and R.~Toledo.
\newblock Lghd: A feature descriptor for matching across non-linear intensity
  variations.
\newblock In {\em Image Processing (ICIP), 2015 IEEE International Conference
  on}, page~5. IEEE, Sep 2015.

\bibitem{aguilera2016learning}
C.~A. Aguilera, F.~J. Aguilera, A.~D. Sappa, and R.~Toledo.
\newblock Learning cross-spectral similarity measures with deep convolutional
  neural networks.
\newblock In {\em Proceedings of the IEEE Conference on Computer Vision and
  Pattern Recognition Workshops}, pages 1--9, 2016.

\bibitem{survey}
L.~G. Brown.
\newblock A survey of image registration techniques.
\newblock {\em ACM computing surveys (CSUR)}, 24(4):325--376, 1992.

\bibitem{multiSpectralSIFT}
M.~Brown and S.~S\"usstrunk.
\newblock Multispectral {SIFT} for scene category recognition.
\newblock In {\em Computer Vision and Pattern Recognition (CVPR11)}, pages
  177--184, Colorado Springs, June 2011.

\bibitem{laplacian}
P.~Burt and E.~Adelson.
\newblock The laplacian pyramid as a compact image code.
\newblock {\em IEEE Transactions on communications}, 31(4):532--540, 1983.

\bibitem{Canny}
J.~Canny.
\newblock A computational approach to edge detection.
\newblock {\em IEEE Transactions on pattern analysis and machine intelligence},
  (6):679--698, 1986.

\bibitem{chen2015sirf}
C.~Chen, Y.~Li, W.~Liu, and J.~Huang.
\newblock Sirf: simultaneous satellite image registration and fusion in a
  unified framework.
\newblock {\em IEEE Transactions on Image Processing}, 24(11):4213--4224, 2015.

\bibitem{dong2015high}
L.~Dong, Q.~Yang, H.~Wu, H.~Xiao, and M.~Xu.
\newblock High quality multi-spectral and panchromatic image fusion
  technologies based on curvelet transform.
\newblock {\em Neurocomputing}, 159:268--274, 2015.

\bibitem{ransac}
M.~A. Fischler and R.~C. Bolles.
\newblock Random sample consensus: a paradigm for model fitting with
  applications to image analysis and automated cartography.
\newblock {\em Communications of the ACM}, 24(6):381--395, 1981.

\bibitem{sobel}
W.~Gao, X.~Zhang, L.~Yang, and H.~Liu.
\newblock An improved sobel edge detection.
\newblock In {\em Computer Science and Information Technology (ICCSIT), 2010
  3rd IEEE International Conference on}, volume~5, pages 67--71. IEEE, 2010.

\bibitem{harris}
C.~Harris and M.~Stephens.
\newblock A combined corner and edge detector.
\newblock In {\em Alvey vision conference}, volume~15, pages 10--5244.
  Manchester, UK, 1988.

\bibitem{IHS}
C.~He, Q.~Liu, H.~Li, and H.~Wang.
\newblock Multimodal medical image fusion based on ihs and pca.
\newblock {\em Procedia Engineering}, 7:280--285, 2010.

\bibitem{huang2014spatial}
B.~Huang, H.~Song, H.~Cui, J.~Peng, and Z.~Xu.
\newblock Spatial and spectral image fusion using sparse matrix factorization.
\newblock {\em IEEE Transactions on Geoscience and Remote Sensing},
  52(3):1693--1704, 2014.

\bibitem{irani1998robust}
M.~Irani and P.~Anandan.
\newblock Robust multi-sensor image alignment.
\newblock In {\em Computer Vision, 1998. Sixth International Conference on},
  pages 959--966. IEEE, 1998.

\bibitem{keller2006multisensor}
Y.~Keller and A.~Averbuch.
\newblock Multisensor image registration via implicit similarity.
\newblock {\em IEEE transactions on pattern analysis and machine intelligence},
  28(5):794--801, 2006.

\bibitem{li1995contour}
H.~Li, B.~Manjunath, and S.~K. Mitra.
\newblock A contour-based approach to multisensor image registration.
\newblock {\em IEEE transactions on image processing}, 4(3):320--334, 1995.

\bibitem{wavelets}
H.~Li, B.~Manjunath, and S.~K. Mitra.
\newblock Multisensor image fusion using the wavelet transform.
\newblock {\em Graphical models and image processing}, 57(3):235--245, 1995.

\bibitem{SIFT}
D.~G. Lowe.
\newblock Distinctive image features from scale-invariant keypoints.
\newblock {\em International Journal of Computer Vision}, 60:91--110, 2004.

\bibitem{lucas1981iterative}
B.~D. Lucas, T.~Kanade, et~al.
\newblock An iterative image registration technique with an application to
  stereo vision.
\newblock 1981.

\bibitem{mutualInformation}
F.~Maes, A.~Collignon, D.~Vandermeulen, G.~Marchal, and P.~Suetens.
\newblock Multimodality image registration by maximization of mutual
  information.
\newblock {\em IEEE transactions on medical imaging}, 16(2):187--198, 1997.

\bibitem{pca}
U.~Patil and U.~Mudengudi.
\newblock Image fusion using hierarchical pca.
\newblock In {\em image Information Processing (ICIIP), 2011 International
  Conference on}, pages 1--6. IEEE, 2011.

\bibitem{correlation}
W.~K. Pratt.
\newblock Correlation techniques of image registration.
\newblock {\em IEEE transactions on Aerospace and Electronic Systems},
  (3):353--358, 1974.

\bibitem{reddy1996fft}
B.~S. Reddy and B.~N. Chatterji.
\newblock An fft-based technique for translation, rotation, and scale-invariant
  image registration.
\newblock {\em IEEE transactions on image processing}, 5(8):1266--1271, 1996.

\bibitem{shen2014multi}
X.~Shen, L.~Xu, Q.~Zhang, and J.~Jia.
\newblock Multi-modal and multi-spectral registration for natural images.
\newblock In {\em European Conference on Computer Vision}, pages 309--324.
  Springer, 2014.

\bibitem{SIFT_Registration}
M.~Subramanyam et~al.
\newblock Automatic feature based image registration using sift algorithm.
\newblock In {\em Computing Communication \& Networking Technologies (ICCCNT),
  2012 Third International Conference on}, pages 1--5. IEEE, 2012.

\bibitem{takacs2010unified}
G.~Takacs, V.~Chandrasekhar, S.~Tsai, D.~Chen, R.~Grzeszczuk, and B.~Girod.
\newblock Unified real-time tracking and recognition with rotation-invariant
  fast features.
\newblock In {\em Computer Vision and Pattern Recognition (CVPR), 2010 IEEE
  Conference on}, pages 934--941. IEEE, 2010.

\bibitem{zitovaSurvey}
B.~Zitova and J.~Flusser.
\newblock Image registration methods: a survey.
\newblock {\em Image and vision computing}, 21(11):977--1000, 2003.

\end{thebibliography}
}

\end{document}